\documentclass{ecai}
\usepackage{times}
\usepackage{graphicx}
\usepackage{latexsym}
\usepackage{float}
\ecaisubmission
\usepackage[utf8]{inputenc} 
\usepackage[T1]{fontenc}    
\usepackage{url}            
\usepackage[hidelinks]{hyperref}
\usepackage{hyperref}       
\usepackage{booktabs}       
\usepackage{amsfonts}       
\usepackage{nicefrac}       
\usepackage{microtype}      
\usepackage{amsmath}
\usepackage{amssymb}
\usepackage{subfigure}
\usepackage{multirow}
\usepackage{tabularx}
\usepackage{graphicx}
\usepackage{times}
\usepackage{graphicx,subfigure,amsmath,bbm,algorithm,algorithmic}
\usepackage[noend]{distribalgo}
\usepackage{xcolor}
\usepackage{caption}
\mathchardef\mhyphen="2D 

\title{Learning Fairness-aware Relational Structures}

\author{Yue Zhang \and Arti Ramesh \institute{SUNY Binghamton,
USA, email: \{yzhan202, artir\}@binghamton.edu} }

\begin{document}

\maketitle
\bibliographystyle{ecai}

\begin{abstract}
The development of fair machine learning models that effectively avert bias and discrimination is an important problem that has garnered attention in recent years. The necessity of encoding complex relational dependencies among the features and variables for competent predictions require the development of fair, yet expressive relational models. In this work, we introduce \textit{Fair-A3SL}, a fairness-aware structure learning algorithm for learning relational structures, which incorporates fairness measures while learning relational graphical model structures. Our approach is versatile in being able to encode a wide range of fairness metrics such as statistical parity difference, overestimation, equalized odds, and equal opportunity, including recently proposed relational fairness measures. While existing approaches employ the fairness measures on pre-determined model structures post prediction, Fair-A3SL directly learns the structure while optimizing for the fairness measures and hence is able to remove any structural bias in the model. We demonstrate the effectiveness of our learned model structures when compared with the state-of-the-art fairness models quantitatively and qualitatively on datasets representing three different modeling scenarios: i) a relational dataset, ii) a recidivism prediction dataset widely used in studying discrimination, and iii) a recommender systems dataset. Our results show that Fair-A3SL can learn fair, yet interpretable and expressive structures capable of making accurate predictions.


\end{abstract}
\section{INTRODUCTION}
\label{sec:related}





The widespread growth and prevalence of machine learning models for crucial decision-making tasks has raised questions on the fairness of the underlying models. Machine learning models have been mostly employed as a black box with little or no transparency or they are too complex to comprehend for non-experts, which further exacerbates this problem. This has led to an increased interest in creating fair machine learning models. The goal of fairness-aware machine learning is to ensure that the decisions made by models do not discriminate against a certain group(s) of individuals \cite{feldman2015certifying,hardt2016equality,boyd2014networked}.


Fairness has been well studied in the social science and policy-making domains \cite{barocas2016big} and is emerging as an important area of research in computer science and specifically, the machine learning community. Most existing work on fairness focus on developing metrics to remove biases after prediction and identifying and removing sensitive attributes \cite{hardt2016equality,kamiran2012decision,pleiss2017fairness} . There is limited existing work on fairness in relational domains.  Farnadi et al.'s \cite{Farnadi2018aaai} work on developing fairness metrics for relational domains and fairness-aware MAP inference for hinge-loss Markov random fields (HL-MRFs) \cite{bach2017hinge} is the first work in this direction. Farnadi et al. \cite{Farnadi2018aaai} note that in many social contexts, discrimination is the result of complex interactions and cannot be described solely in terms of attributes of an individual. While this process is helpful in removing the biases in the inference procedure, it ignores the structural biases in the model structure. This is especially relevant for relational models, where the model structure is instrumental in obtaining the predictions and the biases ingrained in the structure are harder to detect and eliminate. \\
\noindent \textbf{Contributions} \hspace{0.2 cm} In this work, we develop {Fair-A3SL}, a fairness-aware structure learning algorithm for hinge-loss Markov random fields (HL-MRFs). {Fair-A3SL} extends a recently developed deep reinforcement learning-based structure learning algorithm for HL-MRFs, A3SL \cite{ijcai2019838}, to automatically learn \textit{fair} relational graphical model structures.  {Fair-A3SL} has the ability to encode almost all different state-of-the-art widely-used fairness metrics: equalized odds \cite{hardt2016equality}, equal opportunity \cite{hardt2016equality}, statistical parity difference \cite{kamishima2012fairness}, recently developed relational fairness measures of risk difference, risk reward, and relative chance \cite{Farnadi2018aaai}, and fairness measures for collaborative filtering, non-parity and overestimation \cite{yao2017beyond}.  {Fair-A3SL} possesses the ability to encode multiple model-based  and post-processing fairness measures in a single algorithm and can jointly optimize for them to learn a fair model structure. It also offers flexibility in encoding and enforcing these measures through user-defined coefficients that capture the impact of these measures, therefore providing the much needed customizability to enable applicability across multiple domains. 
The added strength of  {Fair-A3SL} arises from its ability to learn interpretable fair structures that do not compromise on performance, further alleviating the problem of opaqueness and lack of interpretability in machine learning models. \emph{To the best of our knowledge, ours is the first approach that directly focuses on learning fair relational model structures from data.}

In our experiments, we demonstrate {Fair-A3SL}'s versatility in being able to encode many different fairness measures and learn fair models for multiple domains. We evaluate the effectiveness of our learned structures in three datasets: i) paper review dataset, a relational dataset used in Farnadi et al. \cite{Farnadi2018aaai} that showcases the ability of our models to learn fair network and collective model structures, ii) Correctional Offender Management Profiling for Alternative Sanctions (COMPAS) dataset, a popular dataset used in many existing fairness work allowing us to compare {Fair-A3SL} with many state-of-the-art fairness models, and iii) MovieLens dataset, a popular dataset used in recommender systems, that enables us to integrate fairness measures used in collaborative filtering in {Fair-A3SL}. {Fair-A3SL} is able to learn structures that eliminate bias at the structure level, requires minimal pre-processing (no other pre-processing other than what is needed for computing the fairness metrics), and can potentially be used easily in sensitive applications to learn interpretable, expressive, and fair model structures that possess good prediction performance for making accurate predictions.

\vspace{-0.2cm}
\section{RELATED WORK}
\label{sec:related}
The state-of-the-art bias mitigation algorithms can be grouped into three categories that include pre-processing, model-based, and post-processing methods. Pre-processing methods work by directly mitigating the bias in the training data itself. Examples of this approach include optimized preprocessing \cite{calmon2017optimized}, which modifies training data features and labels, reweighting \cite{kamiran2012data}, which modifies the weights of different training examples, disparate impact remover \cite{feldman2015certifying}, which edits feature values to improve group fairness, and learning fair representations \cite{zemel2013learning}, which learns fair representations by obfuscating information about protected attributes. 

Model-based methods are used to mitigate bias in classifiers; for example, adversarial debiasing \cite{zhang2018mitigating} uses adversarial techniques to maximize accuracy and reduce evidence of protected attributes in predictions. Prejudice remover \cite{kamishima2012fairness} adds a discrimination-aware regularization term to the learning objective. Meta Fair Classifier \cite{celis2019classification} takes the fairness measure as part of the input and returns a classifier optimized for that metric. Our approach falls in this category.  Existing approaches only learn the parameter values or apply regularization to lessen the effect of sensitive attributes. The fairness measures are not used to directly induce the structure, hence leaving behind some possibility of bias. Our approach differs from existing approaches in that it directly learns the graphical model structure by optimizing for the fairness measures. Thus, our approach is capable of mitigating structural bias in the model, which helps in creating an overall fairer model.

The third class of algorithms focus on post-processing methods to mitigate bias in predictions. For example, reject option classification \cite{kamiran2012decision} changes predictions from a classifier to make them fairer. Equalized odds post-processing \cite{hardt2016equality} modifies the predicted labels using an optimization scheme to make predictions fairer. Calibrated equalized odds post-processing \cite{pleiss2017fairness} optimizes over calibrated classifier score outputs that lead to fair output labels. 


\vspace{-0.3cm}
\section{BACKGROUND FOR FAIR-A3SL}
Before delving into the details of {Fair-A3SL}, we provide necessary background on hinge-loss Markov random fields (HL-MRFs) \cite{bach2017hinge}, the probabilistic programming templating language for encoding them, Probabilistic Soft Logic (PSL) \cite{bach2017hinge}, and a recently developed structure learning algorithm for learning interpretable relational structures in HL-MRFs, asynchronous advantage actor-critic for structure learning (A3SL) \cite{ijcai2019838}.
\vspace{-0.2cm}
\subsection{Hinge-loss Markov Random Fields}
\label{sec:hlmrf}
HL-MRFs are a recently developed scalable class of continuous, conditional graphical models \cite{bach2017hinge}. HL-MRFs can be specified using Probabilistic Soft Logic (PSL) \cite{bach2017hinge}, a first-order logic templating language. In PSL, random variables are represented as logical atoms and weighted rules define dependencies between them of the form:
 $\lambda : P(a) \land Q(a, b) \rightarrow R(b)$,
where \textit{P}, \textit{Q}, and \textit{R} are predicates, \textit{a} and \textit{b} are variables, and $\lambda$ is the weight associated with the rule. The weight of the rule $r$ indicates its importance in the HL-MRF model, which is defined as
\begin{align}
P(\mathit{Y}|\mathit{X}) &\propto \exp \Big (- \sum_{r=1}^M \lambda_r \phi_r(\mathit{Y}, \mathit{X}) \Big ) \nonumber \\
\phi_r(\mathit{Y}, \mathit{X}) &= \left( \max\{l_r(\mathit{Y}, \mathit{X}),0 \}\right)^{\rho_r}  \label{eqn:hl-potential}
\end{align}
\normalsize
where $P(\mathit{Y}|\mathit{X})$ is the probability density function of a subset of logical atoms \textit{Y} given observed logical atoms \textit{X}, $\phi_r(\mathit{Y}, \mathit{X})$ is a hinge-loss potential corresponding to an instantiation of a rule $r$, and is specified by a linear function $l_r$ and optional exponent $\rho_r \in \{1,2\}$. HL-MRFs admit tractable MAP inference regardless of the graph structure of the graphical model, making it feasible to reason over complex user-specified dependencies. This is possible because HL-MRFs operate on continuous random variables and encode dependencies using potential functions that are convex, so MAP inference in these models is always a convex optimization problem. Farnadi et al. \cite{Farnadi2018aaai} extend the MAP inference algorithm to be able to maximize the a-posteriori values of unknown variables subject to fairness guarantees. 


Our approach to learning fair structures focuses on learning logical constructs that particularly bring out the modeling capabilities in HL-MRFs. Below, we provide examples from two datasets we use in our experiments, a relational paper review dataset and a correctional center recidivism prediction dataset: \\
1. Relational Dependencies and Collective Rules: \textit{highQuality(P) $\land$ positiveReviews(R$_1$,P) $\to$ positiveReviews(R$_2$,P)}, which captures if paper \textit{P} is of high quality and reviewer \textit{R}$_1$ gives the paper a positive review, then reviewer \textit{R}$_2$ also gives the paper a positive review. Note that \textit{positiveReviews} is a target predicate and this rule collectively predicts it for both the reviewers. \\
2. Feature Dependencies: \textit{priorFelony(U, $I$) $\land$ africanAmerican(U) $\to$ recidivism(U)}, which captures (unfairly) that if user \textit{U} has committed a prior felony \textit{I} and the race of the user is African American, the user has a higher chance of recidivism. These two features come together to predict recidivism.
\subsection{Asynchronous advantage actor-critic structure learning (A3SL) for HL-MRFs} 

Asynchronous advantage actor-critic structure learning algorithm (A3SL) \cite{ijcai2019838}, a recently developed structure learning algorithm for HL-MRFs, adapts a neural policy gradient algorithm asynchronous advantage actor-critic (A3C) \cite{mnih2016asynchronous} for the structure learning problem. A3SL learns interpretable and expressive structures for HL-MRFs by finding the clause set $C$ and corresponding weight vector $\Lambda$ that maximizes the objective:
$J_\text{A3SL}=L(Y,X)+ \textit{Interpretability Priors}$, where $L(Y,X)$ is the HL-MRF probability density, $logP(Y|X)$, given by Equation \ref{eqn:hl-potential}. $\textit{Interpretability Priors}$ consist of a combination of priors on the total number of clauses, the maximum possible length of a clause, and domain-specific semantic constraints. The inclusion of semantic constraints and a performance-based utility function allows the algorithm to learn structures that are interpretable and data-driven, thus optimizing for both while being able to rectify any domain-specific intuitions that are not true in the data. The objective function $J_\text{A3SL}$ is defined as,
\begin{align}
\label{eqn:a3slobj}
J_\text{A3SL} = &  \Big(L(Y,X) - \alpha_{\textrm{len}}*\frac{1}{|C|}\sum_{c \in C} \textrm{length}(c) \nonumber \\
& -\alpha_{\textrm{num}}* |C| - \alpha_{\textrm{sem}}* \sum_{c \in C}(\textrm{Dist}(c)* \lambda_{c})\Big)
\end{align}
\normalsize
where $\alpha_{\textrm{len}}$, $\alpha_{\textrm{num}}$, and $\alpha_{\textrm{sem}}$ parameters denote the strength of the different constraints, {$\lambda_c$ denotes the weight for PSL clause $c$}, and $\textrm{Dist}(c)$ denotes the deviation of clause $c$ from semantic constraints (discussed more in Section \ref{sec:semantic}). We refer the reader to \cite{ijcai2019838} for additional details.

\section{FAIR-A3SL: FAIRNESS-AWARE STRUCTURE LEARNING FOR HL-MRFS}
\label{sec:problem}

 
In this section, we develop Fair-A3SL by incorporating the different fairness measures in the A3SL problem formulation and objective. We first introduce the {Fair-A3SL} algorithm and then describe all the fairness-related components in the algorithm in detail in the following sections. 

\subsection{Fair-A3SL algorithm}
\label{sec:a3slalg}
Algorithm \ref{alg:faira3sl} gives the Fair-A3SL algorithm. The algorithm follows an actor-critic reinforcement learning setup to learn the clause list $C$ at each step. Our environment consists of predicates for features (denoted by X), target variables (Y), and data corresponding to X and ground truth data for Y.  And each intermediate state $s_t$ at time $t$ comprises of either a partially constructed or a complete set of first order logic clauses, denoted by C. Our action space is defined by all the predicates X, Y, and their negative counterparts, and a special token END. At time $t$, action $a_t$ adds a new predicate to the current clause or chooses to return the clause by adding an END. 

\begin{algorithm}[htb!]
\caption{Fair-A3SL algorithm}
\label{alg:faira3sl}
\small{
\textbf{Input}: A collection of predicates, $X= \{ x_j; j=1,...,m\}$, $Y=\{y_j; j=1,...,n\}$, , Ground truth labels $Y_g$ for $Y$\\ 
Let $C$ = $\{c_0, c_1, .., c_M\}$ denote set of first-order logic clauses, and corresponding weights $\Lambda$  \\
Let $C_{list}$ denote list of $C$ obtained with reward $>$ 0. \\
\textbf{Output}: Optimal $C$ denoted by $C^*$
\begin{distribalgo}[1]
\FUNCTION{{\it $C^*$ = Fair-A3SL($Y$,$X$)} }
\FOR {each thread asynchronously} 
\STATE Construct clause list $C$ under A3SL agent policy
\STATE Initialize weights $\Lambda$ for $C$
\STATE Perform weight learning and update $\Lambda$. 
\STATE Perform fairness-aware inference and get $\hat{Y}$ /* MAP inference with fairness constraints */
\STATE Obtain reward {\it Utility($Y$, $\hat{Y}$)} = $\log P(Y,X)$ - $\alpha*$ \textit{fairness priors}
\STATE Add $C$ to $C_{list}$ 
\STATE Accumulate gradients and update policy and value function parameters according to new state C
\ENDFOR
\STATE $C^{*}$ = optimal $C$ from $C_{list}$
\STATE {\bf return}  $C^{*}$

\ENDFUNC
\end{distribalgo}
%
%
%
}
\end{algorithm}

In the {Fair-A3SL} algorithm, we present two main ways of encoding the fairness measures: i) as MAP inference constraints, and ii) as priors in the objective function. The fairness measures encoded as constraints are integrated as linear inequality constraints in the MAP inference for HL-MRFs; we present more details in Section \ref{sec:constraints}. Step 6 in Algorithm \ref{alg:faira3sl} captures this step, where fairness-aware inference subject to the fairness MAP inference constraints is performed.

To include fairness measures as priors, we turn to the reward/utility function in Step 7 of Algorithm \ref{alg:faira3sl}. The immediate reward $r_t$ is equal to the value of objective function at step $t$ if the clause set construction is complete; $r_t$ equals $0$ otherwise. The cumulative reward $R_t=\sum_{k=0}^\infty\gamma r_{t+k}$ is equal to the value of the objective function, where $\gamma$ is the discount factor, and we set it to 1 in all our experiments. The fairness measures encoded as priors are integrated in the reward utility function, the new utility after incorporating the priors becomes $\textit{Utility}(Y,\hat{Y})=\log P(Y,X)- \alpha*\textit{fairness priors}$, where $\mathit{P(Y|X)}$ is the HL-MRF objective given by Equation \ref{eqn:hl-potential} and $\alpha$ denotes the strength of the fairness prior(s). The algorithm returns the clause list with the best accumulated reward calculated using the utility function as the optimal clause list $C^{*}$.


\vspace{-0.2cm}
\subsection{Fairness aeasures as MAP inference constraints}
\label{sec:constraints}
Here, we discuss how to integrate different fairness measures as MAP inference constraints. First, we start with the assumption that we are given a dataset consisting of $n$ samples $\{(A_i, X_i, Y_i)\}_{i=1}^n$. Here, $A$ denotes one or more sensitive attributes such as gender and race, $X$ denotes other non-sensitive features, and $Y$ denotes the ground-truth labels. We group instances or users based on their sensitive attributes into two groups, protected and unprotected. We then define, $a = \sum_{x \in \textrm{protected group}} \neg \hat{Y}(x)$, $c = \sum_{x \in \textrm{unprotected group}} \neg \hat{Y}(x)$, $g_1 = |\textrm{protected group}|$, $g_2 = |\textrm{unprotected group}|$. $\hat{Y}$ refers to a positive prediction (e.g., acceptance) and $\neg\hat{Y}$ refers to a negative prediction (e.g., denial) from the trained model. The proportions of denial for protected and unprotected groups are $p_1=\frac{a}{g_1}$ and $p_2=\frac{c}{g_2}$, respectively, where $g_1$ and $g_2$ are constants \cite{Farnadi2018aaai,pedreschi2012study}. 

Following Farnadi et al.'s the definition of $\delta$-fairness, the fairness measures can be defined in terms of $p_1$ and $p_2$ as follows, where $0 \leq \delta \leq 1$, \begin{align*}
  & \textrm{Risk difference: RD} = p_1-p_2; \hspace{1 cm} & -\delta \leq \text{RD} \leq \delta   \\
  & \textrm{Risk Ratio: RR} = \frac{p_1}{p_2}; \hspace{1 cm} & 1-\delta \leq \text{RR} \leq 1+\delta\\
  & \textrm{Relative Chance: RC} = \frac{1-p_1}{1-p_2}; \hspace{1 cm} & 1-\delta \leq \text{RC} \leq 1+\delta
\end{align*}
The $\delta$-fairness constraints above translate to \emph{six} linear inequality constraints in the HL-MRF framework. For example, the linear inequality constraints $l_1(Y,X)$ and $l_2(Y,X)$ defined for satisfying the inequality $-\delta \leq$ RD $\leq \delta$ have the forms shown below, where $x_1$,..., $x_{g_1}$ are instances in the protected group, and $x_{g_1+1}, ...,x_{g_1+g_2}$ are instances in the unprotected group, and the total number of instances $n=g_1+g_2$. 
\vspace{-0.4cm}
\begin{align*}
l_1 \Rightarrow \text{RD} \leq \delta 
\Rightarrow \begin{pmatrix}
  g_2 ...g_2,-g_1, ..., -g_1
\end{pmatrix}
*
\begin{pmatrix}
  \hat{Y}(x_1) \\
  \hat{Y}(x_2) \\
  \vdots \\
  \hat{Y}(x_n) \\
\end{pmatrix}
\geq -g_1 g_2 \delta \\
l_2	\Rightarrow \text{RD} \geq -\delta \Rightarrow
\begin{pmatrix}
g_2, ..., g_2, -g_1, ..., -g_1
\end{pmatrix}
*
\begin{pmatrix}
  \hat{Y}(x_1) \\
  \hat{Y}(x_2) \\
  \vdots \\
  \hat{Y}(x_n) \\ 
\end{pmatrix}
\ \leq g_1 g_2 \delta \\
\end{align*}
\vspace{-0.8cm}

Next, we consider a fairness metric for collaborative filtering \cite{yao2017beyond}: non-parity unfairness.  Non-parity unfairness is defined as the absolute difference between the overall predicted average ratings of protected users and those of unprotected users:
\vspace{-0.2cm}
\begin{align}
U_{par}=|E_{\text{protected}}[\hat{Y}]-E_\text{unprotected}[\hat{Y}]| \nonumber \\
E_{\text{protected}}[\hat{Y}] = \frac{1}{g_1}\sum_{\{(i,j)|i \in \text{protected group}\}} \hat{Y}_{i,j} \nonumber \\
E_{\text{unprotected}}[\hat{Y}]=\frac{1}{g_2}\sum_{\{(i,j)|i \in \text{unprotected group}\}} \hat{Y}_{i,j} \nonumber
\end{align} 
where $\hat{Y}$ is the prediction, $g_1$ is the total rating by protected users and $g_2$ the total rating by unprotected users. Below, we demonstrate how to capture non-parity unfairness in {Fair-A3SL} as a MAP inference constraint. We get the corresponding $\delta$-fairness linear inequality constraints $l_3$ and $l_4$ below, where $n$ represents number of users $u$, $m$ represents number of items $v$. 
\begin{align*}
&l_3  \Rightarrow U_{par} \geq -\delta \nonumber \\
& \Rightarrow \begin{pmatrix} 
  g_2 ...g_2,-g_1, ..., -g_1 
\end{pmatrix} 
*\nonumber
\begin{pmatrix} 
  \hat{Y}(u_1, v_1) \nonumber\\
  \hat{Y}(u_1, v_2) \nonumber\\
  \vdots \nonumber\\
  \hat{Y}(u_n, v_m)  \nonumber \\
\end{pmatrix} \geq -g_1g_2\delta \nonumber\\
\end{align*} 
\begin{align*}
 &l_4  \Rightarrow U_{par} \leq \delta \nonumber\\
 & \Rightarrow \begin{pmatrix} \nonumber
  g_2 ...g_2,-g_1, ..., -g_1 \nonumber
\end{pmatrix}\nonumber
*
\begin{pmatrix}\nonumber
  \hat{Y}(u_1, v_1) \nonumber\\
  \hat{Y}(u_1, v_2) \nonumber\\
  \vdots \nonumber\\
  \hat{Y}(u_n, v_m) \nonumber \\
\end{pmatrix} \leq g_1g_2\delta \nonumber
\end{align*} 

The linear form of the constraints is consistent with MAP inference in HL-MRF model; they can be seamlessly solved using a consensus-optimization algorithm based on the alternating direction method of multipliers (ADMM) \cite{boyd2011distributed}. To accomplish this, we extend the consensus optimization algorithm by Bach et al. \cite{bach2017hinge} for MAP inference in HL-MRFs to include above defined fairness linear inequality constraints.
 
Similarly, other fairness measures can also be incorporated in the {Fair-A3SL} framework as constraints. \textit{Statistical Parity Difference} measures the difference of the rate of favorable outcomes received by the unprivileged group to the privileged group \cite{kamishima2012fairness} and \textit{Disparate Impact} measures the ratio of rate of favorable outcome for the unprivileged group to that of the privileged group \cite{feldman2015certifying}. Both these measures are similar to the relative chance (RC) relational measure and can be encoded similarly.

 \vspace{-0.3cm}
\subsection{Fairness measures as objective priors}  
\label{sec:priors}
While certain fairness measures can be modeled as MAP inference constraints in the framework, the post-processing fairness measures can only be modeled as priors in our objective due to the absence of ground truth for target $Y$ at test time as discussed below. 

Equalized Odds Difference \cite{hardt2016equality} measures the difference of false positive rate and true positive rate between unprivileged and privileged groups, which can be defined as $\sum_{y \in \{0,1\}}|Pr(\hat{Y}=1|A=0, Y=y)-Pr(\hat{Y}=1|A=1, Y=y)|$, where $\hat{Y}$ is the predicted value and $Y$ is the ground truth. We cannot directly incorporate this measure as a MAP inference constraint since at test time the true value of $Y$ is not available. This measure and other similar post-processing measures that rely on true ground-truth labels can be encoded as priors in the Fair-A3SL algorithm. We integrate the priors in the objective function, which then is used in computing the agent's rewards in the Fair-A3SL algorithm as discussed in Section \ref{sec:a3slalg}.

Overestimation unfairness measures inconsistency in how much the predictions overestimate the true ratings \cite{yao2017beyond}. This fairness measure is used in the collaborative filtering setting. Following equations give the formula for $U_{\text{over}}$ and the expectation for the protected group $E_{\text{protected}}$. The average for $E_{\text{unprotected}}$ is computed analogously. 
\vspace{-0.2cm}
\begin{align*}
U_{\text{over}}=\frac{1}{m}\sum_{j=1}^{m}\lvert\max(0, E_{\text{protected}}[\hat{Y}]_j-E_{\text{protected}}[Y]_j)  \nonumber \\ 
-\max(0, E_{\text{unprotected}}[\hat{Y}]_j-E_{\text{unprotected}}[Y]_j)\lvert  \\
E_{\text{protected}}[\hat{Y}]_j=\frac{1}{|\{(i,j) | i\in \text{protected}\}|}\sum_{i \in \text{protected}}\hat{Y}_{i,j} 
\end{align*}
Equal Opportunity Difference measures the difference of true positive rates between the unprivileged and the privileged groups \cite{hardt2016equality}. Average Odds Difference \cite{ibmai} measures the average difference of false positive rate and true positive rate between unprivileged and privileged groups. These measures are comparable to the Equalized Odds Difference measure and can be similarly encoded as priors.

\subsection{Domain-specific semantic constraints}
\label{sec:semantic}
An interpretable model lays the foundation for fairness and transparency. In addition to inducing fairness-aware relational structures, we also include semantically meaningful domain constraints that do not contain any structural bias and encourage the algorithm to learn interpretable structures. This is helpful in making the resulting model more appealing to end users. Here, we show how to group predicates and their negative counterparts into two categories, \textit{positive signals} and \textit{negative signals} using the semantic interpretation of the predicate. If the user is unsure about the semantics of any predicate, they can be incorporated in both the categories to avoid any unintentional bias. 
\vspace{-0.2cm}
\begin{table}
\begin{centering}
	\footnotesize
		\caption{{Right reasons identified from domain semantics}}
		    \label{table:rightreasons}
		\begin{tabular}{l}
	\toprule
\textrm{positive signals} $\Rightarrow$ \textrm{any positive signal not already included} \\
\textrm{negative signals} $\Rightarrow$ \textrm{negative signal not already included} \\
\textrm{positive signals} $\land$ $\neg$\textrm{negative signal} $\Rightarrow$ 
\textrm{ positive signal not already included} \\
\textrm{negative signals} $\land \neg$\textrm{ positive signal} $\Rightarrow$ 
\textrm{ negative signal not already included} \\
	\bottomrule
	\end{tabular}
	\end{centering}

\end{table}
\vspace{-0.2cm}

We illustrate this using the COMPAS dataset, one of the datasets widely used in fairness studies and also in our experiments. We capture \textit{positive signals P}=\{\textit{priorFelonHistory, priorMisdemeanorHistory, priorOtherHistory, juvFelonHistory, juvMisdemeanorHistory, juvOtherHistory, priors, felony, recidivism, $\neg$oldAge, longJailDay, $\neg$longJailDay}\} that capture tendency toward recidivism and \textit{negative signals N}=\{\textit{$\neg$felony, $\neg$recidivism, oldAge, longJailDay, $\neg$longJailDay}\} that capture tendency against recidivism. Since at first we are not sure about the effect of \textit{longJailDay} and its negative counterpart on \textit{recidivism} prediction from domain knowledge, we place it in both categories. The domain-specific semantic constraints have the general structure in Table \ref{table:rightreasons}, where positive signals $\subseteq$ $P$, negative signals $\subseteq$ $N$, and any positive signal $\in$ $P$, negative signal $\in$ $N$. We use a distance function, \textit{Dist(c)} to capture if the learned clause structure complies with or deviates from the right reasons identified by the expert: \textit{Dist(c) = 0}, if the clause complies with the right reasons and \textit{Dist(c) = 1}, otherwise. This distance function is then integrated in the objective functions discussed in Section \ref{sec:obj}. If the domain-specific guidance is not readily available for the specific domain, the model is able to work without them as well as they are added only to enhance interpretability when appropriate.


\vspace{-0.2cm}
\subsection{Fair-A3SL objective functions} 
\label{sec:obj}
We present two different objective functions that we use across our three predictive modeling scenarios that demonstrates how a combination of fairness constraints, fairness priors, and semantic constraints can be represented in an objective function. This objective can be easily modified to include/exclude specific fairness/semantic constraints or fairness priors.
\vspace{-0.2cm}
\subsubsection{Fair-A3SL objective for relational models}
In the first objective, we use a combination of fairness measures both encoded as constraints and as priors. Here, we encode the relational fairness measures RR, RC, and RD as MAP inference constraints and the equalized odds difference measure as a prior in the objective along with interpretability priors for the specific domain in question. Equation \ref{eqn:fairobj1} gives the Fair-A3SL objective function corresponding to this combination. We use this objective function in our experiments in Section \ref{sec:relresult} on the relational dataset and in Section \ref{sec:compasresult} on the recidivism prediction dataset.
\begin{align}
\label{eqn:fairobj1}
J_{\textrm{Fair-A3SL}} =&\log P(Y,X)+ \textit{Interpretability Priors} \nonumber \\ 
& + \alpha_\text{odds}*U_\text{odds} \nonumber \nonumber \\
\textrm{s.t.} &-\delta \leq RD \leq \delta \nonumber \\
 & 1-\delta \leq RR \leq 1+\delta \nonumber \\
& 1-\delta \leq RC \leq 1+\delta
\end{align}
where $U_\text{odds}$ refers to the equalized odds difference fairness measure and  $\alpha_\text{odds}$ captures its degree of enforcement.
\vspace{-0.2cm}
\subsubsection{Fair-A3SL objective for recommender systems}
For the recommender systems problem, we turn to the corresponding fairness measures of overestimation and {non-parity}. Equation \ref{eqn:fairobj2} gives the {Fair-A3SL} objective for recommender systems. As is evident from the equation, here again we include a combination of constraints and priors in the objective; we incorporate the {non-parity} fairness measure as a MAP inference constraint ($U_\text{par}$) and {overestimation} as an objective prior ($U_\text{over}$). We use this objective for the experimental results in Section \ref{sec:movieresult}.
\begin{align}
\label{eqn:fairobj2}
J_{\text{Fair-A3SL}}&=\log P(Y,X)+ \alpha_\text{over}*U_\text{over} \nonumber
\\ & s.t. -\delta \leq U_\text{par} \leq \delta 
\end{align}
\vspace{-0.4cm}
\subsection{Highlights of Fair-A3SL} 
Our approach to fairness is versatile in its ability to encode many different fairness measures toward directly learning the graphical model structure. Fair-A3SL  provides the capability of encoding fairness measures as constraints and/or as priors and has minimal pre-processing requirements (only those imposed by the underlying fairness measures). While many existing work indicate the importance of combining fairness measures for practitioners, they also note that there is often a trade-off between various fairness measures and it is challenging to construct a single fairness objective that performs well across different measures \cite{yao2017beyond,farnadi2018fairness}. While this remains true for conflicting measures, {Fair-A3SL} is a step in the right direction, where we present a platform that can incorporate a combination of fairness metrics while simultaneously optimizing for them. In Equations \ref{eqn:fairobj1} and \ref{eqn:fairobj2}, we show some possible combinations and our results indicate Fair-A3SL can indeed optimize for multiple fairness metrics at the same time.  These desirable qualities in Fair-A3SL can potentially help downstream users such as policy makers and decision making organizations (e.g., bank loans, student admissions) to successfully adopt the framework. 

\vspace{-0.2cm}
\section{EXPERIMENTAL EVALUATION}
\label{sec:experiment}
We conduct experiments to evaluate the learned structures quantitatively and qualitatively on three fairness datasets. In our experiments, we illustrate the capability of Fair-A3SL to be able to: i) learn fair network and collective structures that bring out the modeling power of statistical relational models, ii) incorporate a wide range of fairness measures and learn model structures using them, and iii) learn model structures that outperform state-of-the-art fairness models both across performance and fairness metrics and are qualitatively meaningful. The {Fair-A3SL} code and the code for experiments will be made publicly available when the paper is accepted for publication. The best scores and those that are statistically indistinguishable from the best are typed in \textbf{bold} in all the results. All experiments use 5-fold cross-validation.
\vspace{-0.3cm}
\subsection{Results on relational paper review dataset}
\label{sec:relresult}
We first present results on a paper reviewing problem that can potentially be biased by the author's affiliation instead of the quality of the paper. We follow Farnadi et al. \cite{Farnadi2018aaai} to generate a similar dataset to theirs in order to facilitate a direct comparison. 
\begin{table}%
\centering

  	\caption{Generation model of the paper review dataset: \textit{left} shows the joint probability distribution of variables and \textit{right} shows the graphical model. Q: indicates whether or not the paper is high quality; H: indicates whether or not the author is affiliated with a top-rank institute; S: indicates whether or not the author is a student; R1, R2: indicates whether or not the first/second reviewer gives the paper a positive review.}
\parbox{0.4\columnwidth}{
\begin{tabular}{ p{0.2cm}p{0.2cm}p{0.2cm}|p{2.0cm}}
 	Q & H & S & \textrm{P(R1=T | S, Q, H)} \\
 	\midrule
 	F & F & F & 0.15 \\
 	F & F & T & 0.05 \\
 	F & T & F & 0.20 \\
 	F & T & T & 0.15 \\
 	T & F & F & 0.85 \\
 	T & F & T & $\theta_1$ = 0.50 \\
 	T & T & F & 0.85 \\ 
	T & T & T & $\theta_2$ = 0.90 \\
	\end{tabular}
	}  
\qquad
\begin{minipage}[c]{0.2\textwidth}%
\centering
    \includegraphics[width=0.9\textwidth]{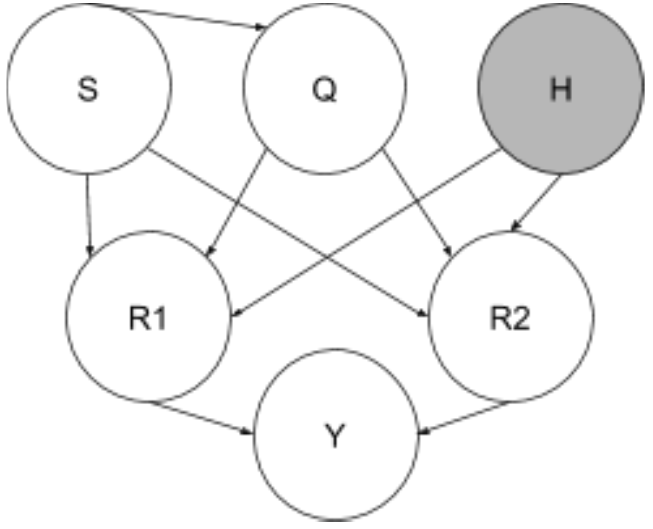}

  \end{minipage}
  	\label{table:gen}

    \label{table:generation_paperReview}
\end{table}
Table \ref{table:generation_paperReview} gives the conditional probability distribution table (\textit{left}) and the Bayesian network (\textit{right}) that we use for generating the data. Two specific scenarios parametrized by P(H) that determine the degree of  discrimination are: i) probability of the paper receiving a favorable rating given the paper is of high quality and the author is not from a top ranked institution ($\theta_1 = P(R_1|Q = T, H = F, S = T)$), and ii) probability of the paper receiving a favorable reviewer rating given the paper is of high quality and the author is from a top ranked institution ($\theta_2 = P(R_1|Q = T,H = T, S=T))$.  We introduce bias in the data when the author is a student (S = T) by setting $\theta_1 = 0.5$ and $\theta_2 = 0.9$. We set $P(R_1|Q = T, H = F, S = F)$ and $P(R_1|Q = T, H = T, S = F)$ to $0.85$. The train and test dataset both contain data generated using the Bayesian network comprising of $100$ papers, $100$ authors, $30$ reviewers, and each paper is reviewed by $2$ random reviewers. 
\begin{table}

\begin{centering}
	\small
		\caption{Fairness-A3SL Model on Paper-Review Dataset}
		\begin{tabular}{l}
	\toprule
	\textbf{PSL Rules Learnt from Fair-A3SL}\\
	\midrule
	Author: $A$; Reviewer : $R_1$, $R_2$; Paper : $P$\\
	\underline{\textbf{Set A. Relational Rules}}: \\
 $\lambda_1$: \textit{submits(A,P) $\land$ student(A) $\land$ positiveReviews($R_1$,P)} \\ \hspace{0.4cm} \textit{$\to$ $\neg$positiveSummary(P)} \\ 
	$\lambda_2$: \textit{acceptable(P) $\land$ positiveReviews($R_1$,P)}  \\ \hspace{0.4cm}  \textit{$\to$ positiveSummary(P)} \\ 
	$\lambda_3$: \textit{$\neg$highQuality(P) $\to$ $\neg$positiveReviews($R_1$,P)} \\ 
$\lambda_4$: \textit{highQuality(P) $\to$ positiveReviews($R_1$,P)} \\ 
	
	\underline{\textbf{Set B. Collective Rules}}:\\
	$\lambda_5$: \textit{highQuality(P) $\land$ positiveReviews($R_1$,P) $\land$ reviews($R_2$,P)} \\ \hspace{0.4cm} \textit{$\to$ positiveReviews($R_2$,P)}\\
	$\lambda_6$: \textit{positiveSummary(P) $\land$ positiveReview($R_1$,P) $\land$ reviews($R_2$,P)} \\ \hspace{0.4cm} \textit{$\to$ positiveReviews($R_2$, P)} \\ 
	
	
	\bottomrule
	\end{tabular}
    \label{table:paperReviewA3SLrules}
	\end{centering}
	\end{table}

Table \ref{table:paperReviewA3SLrules} gives the learnt rules the Fair-A3SL model on the paper review dataset. To enable a comparison with Farnadi et al. \cite{Farnadi2018aaai}, we also enhance A3SL by adding the ability to encode collective rules. Collective rules jointly predict two or more target variables. Note that the learned model structure is expressive, learning different kinds of rules: network, collective, and combination of features.

\begin{table*}
	\begin{centering}
	\footnotesize
	\caption{Comparison of Fair-A3SL with baselines on Area under PR curve and ROC curve}
	\begin{tabular}{ p{3.5cm}p{2.5cm}p{2.5cm}p{2.5cm}}
	\toprule
 	Model & AUC-PR Pro. & AUC-PR Unpro. & AUC-ROC \\
 	\midrule
	Sensitive-PSL & 0.3490$\pm$0.1946 & 0.6112$\pm$0.0566 & 0.8354$\pm$0.0421 \\
	Sensitive-A3SL & 0.3490$\pm$0.1946 & 0.6707$\pm$0.0443 & \textbf{0.8544$\pm$0.0360} \\
	Fair-PSL (\cite{Farnadi2018aaai}) & 0.4332$\pm$0.1104 & 0.6009$\pm$0.0463 & 0.7887$\pm$0.0306 \\
	$\textrm{Fair-A3SL}_1$ & 0.3490$\pm$0.1946 & 0.6112$\pm$0.0566 & 0.8037$\pm$0.0491 \\
	$\textrm{Fair-A3SL}_2$ & \textbf{0.6208$\pm$0.2981} & \textbf{0.7279$\pm$0.0322} & 0.8118$\pm$0.0076 \\
	$\textrm{Fair-A3SL}_3$ & \textbf{0.6208$\pm$0.2981} & 0.5469$\pm$0.1486 & 0.7396$\pm$0.0005 \\
 	\bottomrule
	\end{tabular}
    \label{table:results_paperReview1}
	\end{centering}
	\begin{centering}
	\footnotesize
	\caption{Comparison of Fair-A3SL with baselines on fairness measures}
	\begin{tabular}{ p{2cm}p{2.2cm}p{2.2cm}p{2.2cm}p{2.2cm}p{2.2cm}}
	\toprule
 	Model & RD & RR & RC & Equal Odds Pos. & Equal Odds Neg.\\
 	\midrule
	Sensitive-PSL & 0.2766$\pm$0.0768 & 0.2090$\pm$0.0935 & 1.4344$\pm$0.1543 & 0.5227$\pm$0.1799 & 0.1429$\pm$0.0389\\
	Sensitive-A3SL & 0.0661$\pm$0.0591 & 0.8723$\pm$0.1479 & 1.1094$\pm$0.1202 & 0.1550$\pm$0.1156 & 0.0317$\pm$0.0056\\
	Fair-PSL \cite{Farnadi2018aaai} & \textbf{0.0005$\pm$0.0004} & \textbf{0.9980$\pm$0.0019} & \textbf{1.0007$\pm$0.0007} & 0.1346$\pm$0.0234 & 0.0829$\pm$0.0550\\
	$\textrm{Fair-A3SL}_1$ & \textbf{0.0002$\pm$3.7e-5} & \textbf{1.0007$\pm$0.0002} & \textbf{0.9996$\pm$4.9e-5} & 0.1968$\pm$0.1899 & 0.1286$\pm$0.0548\\
	$\textrm{Fair-A3SL}_2$ & \textbf{0.0059$\pm$0.0005} & \textbf{1.0009$\pm$0.0115} & \textbf{0.9989$\pm$0.0122} & \textbf{0.0096$\pm$0.0066} & \textbf{0.0093$\pm$0.0054}\\
	$\textrm{Fair-A3SL}_3$ & \textbf{7.9e-5$\pm$1.8e-5} & \textbf{0.9999$\pm$4.9e-5} & \textbf{1.1651$\pm$0.3413} & \textbf{0.0001$\pm$1.2e-5} & \textbf{6.6e-5$\pm$3.2e-5} \\
 	\bottomrule
	\end{tabular}
    \label{table:results_paperReview2}
	\end{centering}
%
	\end{table*}

We compare Fair-A3SL with the following state-of-the-art {baselines}: i) Fair-PSL \cite{Farnadi2018aaai}, manually-defined PSL rules with fairness constraints in inference, ii) Sensitive-PSL, manually-defined PSL rules with \textit{no} fairness constraints, and iii) Sensitive-A3SL \cite{ijcai2019838}, a model structure learned using A3SL with \textit{no} fairness constraints or priors. Additionally, we experiment with three versions of Fair-A3SL that use different combinations of fairness measures. $\textrm{Fair-A3SL}_1$ includes fairness constraints without equalized odds priors. $\textrm{Fair-A3SL}_2$ includes fairness constraints along with equalized odds priors with $\alpha_{odds} = 0.1$. $\textrm{Fair-A3SL}_3$ includes fairness constraints along with equalized odds with $\alpha_\text{odds} = 0.5$. We set $\delta$-\textrm{fairness}=$0.1$ for all fairness inference inequality constraints. The AUC-ROC values from the Sensitive-A3SL model can be considered an upper bound, as it is a purely data-driven model.

Our specific focus is on the prediction performance for protected/unprotected groups, especially for predicting a positive outcome in both these groups (Table \ref{table:results_paperReview1}). We report area under the AUC-PR curve for the positive class (\textit{positiveSummary}). From the table, we can see that all A3SL versions outperform the human expert counterparts (Sensitive-A3SL vs. Sensitive-PSL, Fair-A3SL versions vs. Fair-PSL). We can see that the Fair-PSL model even when the fairness measures are included in the inference only achieves a prediction performance of $\sim0.4$, while the Fair-A3SL models achieve $>0.6$ for the protected group. The Fair-A3SL models also improve the prediction performance of the unprotected groups when compared to the Fair-PSL model. The combined AUC-ROC value for the Fair-A3SL models is also closer to the models that include sensitive attributes (Sensitive-PSL and Sensitive-A3SL). Similarly, all the Fair-A3SL models achieve better or comparable performance across all fairness metrics (RD, RR, RC, Equalized Odds Positive and Negative) when compared with Fair-PSL with manually defined rules (Table \ref{table:results_paperReview2}). Particularly, for the equalized odds measures, Fair-A3SL models clearly outperform Fair-PSL. We also observe that we get better results for the equalized odds fairness measure when we increase the value of $\alpha_\text{odds}$. Thus, Fair-A3SL is able to achieve fairness without compromising on performance.

\vspace{-0.3cm}
\subsection{Results on COMPAS dataset}
\label{sec:compasresult}
The Correctional Offender Management Profiling for Alternative Sanctions (COMPAS) tool produces a risk score that predicts a person's likelihood of committing a crime in the next two years \cite{compas}. The output is a score between 1 to 10 that maps to low, medium, or high. We collapse this to a binary prediction: a score of 0 corresponds to a prediction of low risk according to COMPAS, while a score of 1 indicates high or medium risk. The dataset also contains information on recidivism for each person over the next two years, which we use as ground truth. Existing work shows that the COMPAS risk scores discriminate against black defendants, who were predicted to be far more likely than white defendants to be incorrectly judged to be at a higher risk of recidivism, while white defendants were more likely than black defendants to be incorrectly flagged as low risk \cite{compas,dressel2018accuracy}.  

Table \ref{table:compassensitiveA3SL}  gives the Sensitive-A3SL model. We can see that the model combines other recidivism signals of having committed prior felonies (\textit{priors} and \textit{priorFelony)} with the race attribute (\textit{africanAmerican}), indicating how the race attribute and combinations with it are predictive of recidivism and are a natural albeit unfair and discriminatory choice for models that are solely performance driven. The rules learned by the Fair-A3SL model are given in Table \ref{table:compasA3SLrules}. Parameter $U$ represents user, $I_i$ represents a felony instance. For example,  \textit{priorFelonHistory(U,$I_1$)} can be grounded with multiple historical felony instances $I_1$ for each user $U$. 
\vspace{-0.35cm}
\begin{table}
	 \small
	 	\caption{Representative rules from Sensitive-A3SL model}
	 \begin{tabular}{l}
	\toprule
	\textbf{Sensitive-A3SL Recidivism Model}\\
	\midrule
	$U:$ users; $I_i:$ felony instances.\\
	\midrule
	{\textit{priors(U, $I_4$) $\land$ africanAmerican(U) $\to$ recidivism(U)}} \\
	{\textit{priorFelony(U, $I_5$) $\land$ africanAmerican(U) $\to$ recidivism(U)} }\\
	{\textit{$\neg$oldAge(U) $\land$ africanAmerican(U) $\to$ recidivism(U)}}\\
	{\textit{africanAmerican(U) $\to$ recidivism(U)}} \\
	\bottomrule
	\end{tabular}
    \label{table:compassensitiveA3SL}
	\small
		\caption{Rules from Fair-A3SL model}
	\begin{tabular}{l}
	\toprule
	\textbf{Fairness-A3SL Recidivism Model}\\
	\midrule
	$U:$ users; $I_i:$ felony instances.\\
	\midrule
	\underline{\textbf{Set A. Combining Local Features}}: \\
	$\lambda_1$: \textit{oldAge(U) $\to$ $\neg$ recidivism(U)} \\
	 $\lambda_2$: \textit{$\neg$oldAge(U) $\land$ longJailDay(U) $\to$ recidivism(U)} \\
	 $\lambda_3$: \textit{$\neg$longJailDay(U) $\to$ recidivism(U)} \\

	\underline{\textbf{Set B. Combining Jail History Features}}: \\	
	$\lambda_4$: \textit{priorFelonHistory(U, $I_1$) $\to$ recidivism(U)} \\
	$\lambda_5$: \textit{priorMisdemeanorHistory(U, $I_2$) $\to$ recidivism(U)} \\
	$\lambda_6$: \textit{juvenileOtherHistory(U, $I_3$) $\to$ recidivism(U)} \\
	$\lambda_7$: \textit{priors(U, $I_4$) $\to$ recidivism(U)} \\
	
	\underline{\textbf{Set C. Prior Rule}}:\\
	 $\lambda_8$: \textit{user(U) $\to \neg$ recidivism(U)}\\	
	
	
	\bottomrule
	\end{tabular}
    \label{table:compasA3SLrules}
\end{table}
\vspace{-0.4cm}
\begin{table*}[ht!]
	\begin{centering}
	\footnotesize
		\caption{AUC-PR curve and ROC values for state-of-the-art fairness models and Fair-A3SL for COMPAS dataset.}
	\begin{tabular}{ p{4cm}p{2.5cm}p{2.5cm}p{2.5cm}}
	\toprule
 	Model & AUC-PR Pro. & AUC-PR Unpro. & AUC-ROC \\
 	\midrule
	COMPAS Scores \cite{compas} & 0.6168$\pm$0.0177 & 0.5118$\pm$0.0173 & 0.6530$\pm$0.0158 \\
	Line-FERM \cite{donini2018empirical} & 0.6281$\pm$0.0077 & 0.5103$\pm$0.0167 & 0.6482$\pm$0.0145 \\
	Calibrated Equalized Odds \cite{pleiss2017fairness} & 0.7030$\pm$0.0371 & 0.3882$\pm$0.0096 & 0.6553$\pm$0.0207 \\
	Prejudice Remover \cite{kamishima2012fairness}& 0.6801$\pm$0.0274 & 0.5494$\pm$0.0259 & 0.6859$\pm$0.0040 \\
	Optimized Preprocessing \cite{calmon2017optimized} & 0.7039$\pm$0.0191 & 0.5903$\pm$0.0399 & \textbf{0.7131$\pm$0.0123} \\
	Adversarial Debiasing \cite{zhang2018mitigating} & 0.6654$\pm$0.0334 & 0.5174$\pm$0.0356 & 0.6545$\pm$0.0241 \\
	\textbf{Fair-A3SL} (our approach)& \textbf{0.7262$\pm$0.0165} & \textbf{0.6080$\pm$0.0229} & \textbf{0.7103$\pm$0.0109} \\
 	\bottomrule
	\end{tabular}
    \label{table:accuracy_results_compas}
	\end{centering}
	\begin{centering}
	\footnotesize
	\caption{Comparison of performance of Fair-A3SL with state-of-the-art fairness models on different fairness metrics for COMPAS dataset}
	\begin{tabular}{ p{4cm}p{1.7cm}p{1.7cm}p{1.7cm}p{2cm}p{2.2cm}}
	\toprule
 	Model & RD & RR & RC & Equal Odds Pos. & Equal Odds Neg.\\
 	\midrule
	COMPAS Scores \cite{compas} & 0.2632$\pm$0.0228 & 1.8170$\pm$0.1165 & 0.6106$\pm$0.0232 & 0.2261$\pm$0.0284 & 0.2285$\pm$0.0187\\
	Line-FERM \cite{donini2018empirical} & 0.1450$\pm$0.0647 & 1.5485$\pm$0.2704 & 0.7936$\pm$0.0961 & 0.1147$\pm$0.0774 & 0.1063$\pm$0.0609 \\
	Calibrated Equalized Odds \cite{pleiss2017fairness} & 0.1350$\pm$0.0145 & 1.3480$\pm$0.0515 & 0.7986$\pm$0.0322 & 0.1946$\pm$0.0180 & 0.0698$\pm$0.0160\\
	Prejudice Remover \cite{kamishima2012fairness}& 0.0541$\pm$0.0089 & 1.1438$\pm$0.0306 & 0.9125$\pm$0.0124 & 0.0772$\pm$0.0194 & 0.0583$\pm$0.0118\\
	Optimized Pre-processing \cite{calmon2017optimized} & 0.0517$\pm$0.0102 & 1.1218$\pm$0.0259 & 0.9099$\pm$0.0168 & 0.0325$\pm$0.0178 & 0.0361$\pm$0.0088\\
	Adversarial Debiasing \cite{zhang2018mitigating} & 0.0511$\pm$0.0096 & 1.1176$\pm$0.0243 & 0.9094$\pm$0.0161 & 0.0539$\pm$0.0089 & 0.0307$\pm$0.0089\\
	\textbf{Fair-A3SL} (our approach)& \textbf{0.0035$\pm$0.0003} & \textbf{1.0047$\pm$0.0006} & \textbf{0.9851$\pm$0.0030} & \textbf{0.0039$\pm$0.0051} & \textbf{0.0160$\pm$0.0105} \\
 	\bottomrule
	\end{tabular}
    \label{table:fairness_results_compas}
	\end{centering}
\end{table*}
Fair-A3SL's transparency, interpretability, expressibility, along with fairness, makes it an ideal candidate for automatically learning prediction models for sensitive domains. 

We compare Fair-A3SL with recently developed state-of-the-art fairness models: i) Calibrated Equalized Odds \cite{pleiss2017fairness}, ii) Prejudice Remover \cite{kamishima2012fairness}, iii) Optimized Pre-processing \cite{calmon2017optimized}, iv) Adversarial Debiasing \cite{zhang2018mitigating}, and v) Line-FERM \cite{donini2018empirical}, where Calibrated Equalized Odds, Prejudice Remover, and Optimized Preprocessing use logistic regression as the backend model; Adversarial Debiasing uses a deep learning neural network model; and FERM uses SVM as the underlying model. Table \ref{table:accuracy_results_compas} gives the 5-fold cross-validation results and shows that Fair-A3SL is able to achieve a better prediction performance for both the protected and unprotected groups, individually (AUC-PR for protected and unprotected groups) and combined (AUC-ROC).  We use the IBM AI Fairness 360 tool \cite{ibmai} for running the existing state-of-the-art models. We also demonstrate that our learned model outperforms the state-of-the-art fairness models in the fairness metrics as well, achieving the best scores across all metrics (Table \ref{table:fairness_results_compas}). 


\vspace{-0.4cm}
\subsection{Results on Movielens dataset}
\label{sec:movieresult}
In the third experiment, we consider another important domain for fairness, recommender systems. To evaluate the effectiveness of {Fair-A3SL} in recommender systems, we use the \textit{MovieLens} $100k$ dataset. It consists of ratings from 1 to 5 by 943 users for 1682 movies. The users are annotated with demographic variables such as gender, and the movies are each annotated with a set of genres. For convenience, we convert the ratings to range between values 0 and 1. From Table \ref{table:movie_statics}, we can see that women rate musical and romance films higher and more frequently than men. Men rate Sci-Fi and crime films higher and more frequently than women. Women and men both give action films an almost equal rating, but men rate these films more frequently.

\begin{table}

	\begin{centering}
	\scriptsize
		\caption{Gender-based statistics of movie genres in MovieLens data.}
	\begin{tabular}{ p{2.3cm}p{0.8cm}p{0.7cm}p{0.8cm}p{0.7cm}p{0.7cm}}
	\toprule
 	{} & Romance & Action & Sci-Fi & Musical & Crime \\
 	\midrule
	Count & 14202 & 19141 & 9577 & 3765 & 5835 \\
	Avg Count per Female & 24.74 & 23.13 & 11.57 & 7.32 & 7.41 \\
	Avg Count per Male & 20.43 & 31.11 & 15.64 & 6.30 & 9.67 \\
	Avg Rating by Female & 0.73 & 0.70 & 0.70 & 0.73 & 0.71 \\
	Avg Rating by Male & 0.72 & 0.70 & 0.71 & 0.69 & 0.73 \\
	\bottomrule
	\end{tabular}
    \label{table:movie_statics}
	\end{centering}
	 \small
	 \caption{Rules from Fair-A3SL model for MovieLens data}
	\begin{tabular}{l}
	\toprule
	\textbf{Fair-A3SL Recommender Model}\\ 
	\midrule
	$U_i:$ users; $I_i:$ items.\\
	\midrule
	{$\lambda_1$: \textit{$\text{rating}_{MF}(U,I)$ $\to$ rating(U,I)}} \\
	{$\lambda_2$: \textit{avgUserRating(U) $\land$ reviews(U,I) $\to$ rating(U,I)}}\\
	{$\lambda_3$: \textit{itemPearsonSim(I, $I_2$) $\land$ rating$_{MF}$(U,I) $\land$ rating(U,I)}}\\ 
	{\hspace{0.4cm} \textit{$\to$ rating(U,$I_2$)}} \\
	{$\lambda_4$: \textit{userPearsonSim(U, $U_2$) $\land$ rating($U_2$,I) $\land$ avgUserRating(U)}} \\
	{\hspace{0.4cm} \textit{$\to$ avgItemRating(I)}} \\
	\bottomrule
	\end{tabular}
    \label{table:recommenderfairA3SL}
\end{table} 

\begin{table}
	\begin{centering}
	\scriptsize
	\caption{Mean square errors (MSE) results on state-of-the-art fairness models and Fair-A3SL on MovieLens dataset}
	\begin{tabular}{ p{1.45cm}p{1.85cm}p{1.85cm}p{1.85cm}}
	\toprule
 	Model & Err Pro. & Err Unpro. & Error \\
 	\midrule
	\text{HyPER \cite{kouki2015hyper}} & 0.04530$\pm$0.00212 & 0.03887$\pm$7.4e-5 & 0.04043$\pm$0.00046 \\
	\text{MF} \cite{koren2009matrix} & 0.03909$\pm$0.00233 & 0.03258$\pm$0.00014 & 0.03415$\pm$0.00067 \\
	\text{Fair-HyPER}\cite{farnadi2018fairness} & 0.03947$\pm$0.00222 & 0.03297$\pm$0.00015 & 0.03455$\pm$0.00065 \\
	\text{Fair-MF} \cite{yao2017beyond} & 0.03942$\pm$0.00215 & 0.03249$\pm$0.00015 & 0.03415$\pm$0.00063 \\
	\textbf{Fair-A3SL} & \textbf{0.03779$\pm$0.00203} & \textbf{0.03189$\pm$0.00025} & \textbf{0.03331$\pm$0.00068} \\
 	\bottomrule
	\end{tabular}
    \label{table:accuracy_results_movie}
	\end{centering}

	\begin{centering}
	\scriptsize
	\caption{Overall fairness measurements of state-of-the-art fairness models and Fair-A3SL on MovieLens dataset}
	\begin{tabular}{ p{3.3cm}p{2.0cm}p{2.0cm}}
	\toprule
 	Model & Non-Parity & Overestimation \\
 	\midrule
	\text{HyPER \cite{kouki2015hyper}}  & 0.00424$\pm$0.00033  & \textbf{0.0349$\pm$0.00338}\\
	\text{MF} \cite{koren2009matrix} & 0.00473$\pm$0.0005 & 0.06294$\pm$0.00475  \\
	\text{Fair-HyPER} \cite{farnadi2018fairness}  & 0.00465$\pm$0.00037 & 0.05346$\pm$0.00380 \\
	\text{Fair-MF} \cite{yao2017beyond}  & 0.00076$\pm$0.00055  & 0.06101$\pm$0.00402\\
	\textbf{Fair-A3SL}  & \textbf{9.2e-5$\pm$5.9e-5}  & {0.05914$\pm$0.00307}\\
 	\bottomrule
	\end{tabular}
    \label{table:fairness_results_movie}
	\end{centering}
\end{table}

Following Kouki et al. \cite{kouki2015hyper}, we extract features that combines multiple different sources of information, including similarity between pairs of users (\textit{userPearsonSim($U,U_2$)}), similarity between items (\textit{itemPearsonSim}($I,I_2$)), average rating with respect to users and items to serve as priors (\textit{avgUserRating}($U$) and \textit{avgItemRating}($I$)), and leveraging predictions from existing recommendation algorithms as a feature (\textit{rating$_{MF}(U,I)$}) to enable an appropriate comparison. Table \ref{table:recommenderfairA3SL} gives the rules learned by Fair-A3SL. 

We compare our approach to the state-of-the-art recommender systems baseline models: i) HyPER \cite{kouki2015hyper}, which is a PSL model and includes hybrid recommender systems feature,; ii) matrix factorization based collaborative filtering model \cite{koren2009matrix}, iii) Fair-HyPER \cite{farnadi2018fairness}, which defines additional latent variable rules to abstract the rating of unprotected and protected groups in order to ensure there is no overestimation unfairness, iv) baseline model Fair-MF \cite{yao2017beyond}, which considers overestimation and non-parity unfairness as regularization terms. Table \ref{table:accuracy_results_movie} shows Fair-A3SL achieves the best overall performance for both the protected and unprotected groups. Table \ref{table:fairness_results_movie} shows that our Fair-A3SL model gets a comparable value in the overestimation unfairness measure, and the best value in the non-parity fairness measure. The model learned by Fair-A3SL achieves comparable performance to Fair-HyPER even without the inclusion of carefully designed latent variables that provide additional complexity. 

%
%
%


\vspace{-0.3cm}
\section{CONCLUSION}
\label{sec:conclusion}
In this work, we developed Fair-A3SL, a general purpose fair structure learning algorithm for HL-MRFs and demonstrated that it learns fair, semantically interpretable, and expressive relational structures while achieving good prediction performance. {Fair-A3SL} is capable of encoding various different measures of fairness both as constraints and priors and we demonstrate its effectiveness across three different domains and modeling scenarios. Further, {Fair-A3SL} has minimal pre-processing requirements (only those posed by the underlying fairness measures) and can seamlessly be utilized to learn models for any sensitive prediction problem including those that require complex relational structures. Fair-A3SL's joint qualities of fairness, interpretability, and performance make it lucrative for many downstream applications (e.g., bank loans, student admissions) to adopt it.


%

%
\bibliographystyle{ecai}

\begin{thebibliography}{10}

\bibitem{ibmai}
{IBM} {AI} fairness 360 open source toolkit.
\newblock \url{https://aif360.mybluemix.net/}.

\bibitem{bach2017hinge}
Stephen~H Bach, Matthias Broecheler, Bert Huang, and Lise Getoor, `Hinge-loss
  markov random fields and probabilistic soft logic', {\em Journal of Machine
  Learning Research (JMLR)}, {\bf 18}(109),  1--67, (2017).

\bibitem{barocas2016big}
Solon Barocas and Andrew~D Selbst, `Big data's disparate impact', {\em
  California Law Review}, {\bf 104},  671, (2016).

\bibitem{boyd2014networked}
Danah Boyd, Karen Levy, and Alice Marwick, `The networked nature of algorithmic
  discrimination', {\em Data and Discrimination: Collected Essays. Open
  Technology Institute}, (2014).

\bibitem{boyd2011distributed}
Stephen Boyd, Neal Parikh, Eric Chu, Borja Peleato, Jonathan Eckstein, et~al.,
  `Distributed optimization and statistical learning via the alternating
  direction method of multipliers', {\em Foundations and
  Trends{\textregistered} in Machine learning},  1--122, (2011).

\bibitem{calmon2017optimized}
Flavio Calmon, Dennis Wei, Bhanukiran Vinzamuri, Karthikeyan~Natesan
  Ramamurthy, and Kush~R Varshney, `Optimized pre-processing for discrimination
  prevention', in {\em Proceedings of the Conference on Advances in Neural
  Information Processing Systems (NIPS)}, (2017).

\bibitem{celis2019classification}
L~Elisa Celis, Lingxiao Huang, Vijay Keswani, and Nisheeth~K Vishnoi,
  `Classification with fairness constraints: A meta-algorithm with provable
  guarantees', in {\em Proceedings of the Conference on Fairness,
  Accountability, and Transparency (FAT*)}, (2019).

\bibitem{donini2018empirical}
Michele Donini, Luca Oneto, Shai Ben-David, John~S Shawe-Taylor, and
  Massimiliano Pontil, `Empirical risk minimization under fairness
  constraints', in {\em Proceedings of the Conference on Advances in Neural
  Information Processing Systems (NIPS)}, (2018).

\bibitem{dressel2018accuracy}
Julia Dressel and Hany Farid, `The accuracy, fairness, and limits of predicting
  recidivism', {\em Science advances}, {\bf 4}(1),  eaao5580, (2018).

\bibitem{Farnadi2018aaai}
Golnoosh Farnadi, Behrouz Babaki, and Lise Getoor, `Fairness in relational
  domains', in {\em Proceedings of the AAAI/ACM Conference on AI, Ethics, and
  Society (AIES)}, (2018).

\bibitem{farnadi2018fairness}
Golnoosh Farnadi, Pigi Kouki, Spencer~K Thompson, Sriram Srinivasan, and Lise
  Getoor, `A fairness-aware hybrid recommender system', in {\em RecSys Workshop
  on FATREC}, (2018).

\bibitem{feldman2015certifying}
Michael Feldman, Sorelle~A Friedler, John Moeller, Carlos Scheidegger, and
  Suresh Venkatasubramanian, `Certifying and removing disparate impact', in
  {\em Proceedings of the International Conference on Knowledge Discovery and
  Data Mining (KDD)}, (2015).

\bibitem{hardt2016equality}
Moritz Hardt, Eric Price, Nati Srebro, et~al., `Equality of opportunity in
  supervised learning', in {\em Proceedings of the Conference on Advances in
  neural information processing systems (NIPS)}, (2016).

\bibitem{kamiran2012data}
Faisal Kamiran and Toon Calders, `Data preprocessing techniques for
  classification without discrimination', {\em Knowledge and Information
  Systems (KAIS)},  1--33, (2012).

\bibitem{kamiran2012decision}
Faisal Kamiran, Asim Karim, and Xiangliang Zhang, `Decision theory for
  discrimination-aware classification', in {\em Proceedings of the
  International Conference on Data Mining (ICDM)}, (2012).

\bibitem{kamishima2012fairness}
Toshihiro Kamishima, Shotaro Akaho, Hideki Asoh, and Jun Sakuma,
  `Fairness-aware classifier with prejudice remover regularizer', in {\em
  Proceedings of the Joint European Conference on Machine Learning and
  Knowledge Discovery in Databases}, (2012).

\bibitem{koren2009matrix}
Yehuda Koren, Robert Bell, and Chris Volinsky, `Matrix factorization techniques
  for recommender systems', {\em Computer}, (8),  30--37, (2009).

\bibitem{kouki2015hyper}
Pigi Kouki, Shobeir Fakhraei, James Foulds, Magdalini Eirinaki, and Lise
  Getoor, `Hyper: A flexible and extensible probabilistic framework for hybrid
  recommender systems', in {\em Proceedings of the ACM Conference on
  Recommender Systems (RecSys)}, (2015).

\bibitem{compas}
Jeff Larson, Surya Mattu, Lauren Kirchner, and Julia Angwin, `How we analyzed
  the {COMPAS} recidivism algorithm', in {\em ProPublica}, (2016).

\bibitem{mnih2016asynchronous}
Volodymyr Mnih, Adria~Puigdomenech Badia, Mehdi Mirza, Alex Graves, Timothy
  Lillicrap, Tim Harley, David Silver, and Koray Kavukcuoglu, `Asynchronous
  methods for deep reinforcement learning', in {\em Proceedings of the
  International Conference on Machine Learning (ICML)}, (2016).

\bibitem{pedreschi2012study}
Dino Pedreschi, Salvatore Ruggieri, and Franco Turini, `A study of top-k
  measures for discrimination discovery', in {\em Proceedings of the Annual ACM
  Symposium on Applied Computing}, (2012).

\bibitem{pleiss2017fairness}
Geoff Pleiss, Manish Raghavan, Felix Wu, Jon Kleinberg, and Kilian~Q
  Weinberger, `On fairness and calibration', in {\em Proceedings of the
  Conference on Advances in Neural Information Processing Systems (NIPS)},
  (2017).

\bibitem{yao2017beyond}
Sirui Yao and Bert Huang, `Beyond parity: Fairness objectives for collaborative
  filtering', in {\em Proceedings of the Conference on Advances in Neural
  Information Processing Systems (NIPS)}, (2017).

\bibitem{zemel2013learning}
Rich Zemel, Yu~Wu, Kevin Swersky, Toni Pitassi, and Cynthia Dwork, `Learning
  fair representations', in {\em Proceedings of the International Conference on
  Machine Learning (ICML)}, (2013).

\bibitem{zhang2018mitigating}
Brian~Hu Zhang, Blake Lemoine, and Margaret Mitchell, `Mitigating unwanted
  biases with adversarial learning', in {\em Proceedings of the AAAI/ACM
  Conference on AI, Ethics, and Society (AIES)}, (2018).

\bibitem{ijcai2019838}
Yue Zhang and Arti Ramesh, `Learning interpretable relational structures of
  hinge-loss markov random fields', in {\em Proceedings of the International
  Joint Conference on Artificial Intelligence (IJCAI)}, (2019).

\end{thebibliography}

\end{document}